\documentclass[11pt]{article}

\usepackage[margin=1.0in]{geometry}
\usepackage[T1]{fontenc}
\usepackage[utf8]{inputenc}
\usepackage{times}
\usepackage{amsmath,amssymb,amsthm,mathtools}
\usepackage{booktabs}
\usepackage{multirow}
\usepackage{adjustbox}
\usepackage{array}
\usepackage{graphicx}
\usepackage{subcaption}
\usepackage{float}
\usepackage[numbers,compress]{natbib}
\usepackage[colorlinks=true,linkcolor=blue!60!black,citecolor=blue!60!black,urlcolor=blue!60!black]{hyperref}
\usepackage{url}
\usepackage{parskip}
\usepackage{enumitem}
\usepackage{xcolor}
\usepackage{titlesec}
\usepackage{setspace}

\titleformat{\section}{\large\bfseries}{\thesection}{1em}{}
\titleformat{\subsection}{\normalsize\bfseries}{\thesubsection}{1em}{}
\titleformat{\paragraph}[runin]{\normalsize\bfseries}{}{0em}{}[.\quad]

\newcommand{\calC}{\mathcal{C}}
\newcommand{\calR}{\mathcal{R}}
\newcommand{\Var}{\mathrm{Var}}
\newcommand{\Med}{\mathrm{median}}
\newcommand{\clamp}{\mathrm{clamp}}

\title{\textbf{How Sparsity Allocation Shapes Label-Free Post-Pruning Recoverability}}

\author{
  Qishi Zhan$^{1}$ \quad Minxuan Hu$^{2}$ \quad Liang He$^{3}$ \\[4pt]
  $^{1}$Marquette University \quad
  $^{2}$Cornell University \quad
  $^{3}$Tongji University \\[2pt]
  \texttt{qishi.zhan@marquette.edu}
}

\date{}

\begin{document}

\maketitle

\begin{abstract}
\noindent
Unstructured magnitude pruning at high sparsity ratios can reduce a neural
network's accuracy to near-random performance, yet retraining with labeled
data is often infeasible in practical deployment settings. Recent work on
Adaptive Signal Resuscitation (ASR) shows that label-free, gradient-free
channelwise repair can recover substantial accuracy from collapsed sparse
models~\citep{zhan2026asr}. Building on this repair operator as a fixed
backend, we ask a different question: how does the choice of upstream sparsity
allocation affect the amount of activation signal that remains recoverable
after pruning? We evaluate ERK and LAMP allocations under the same ASR repair
protocol across CIFAR-10, CIFAR-100, and Imagenette with ResNet-18, ResNet-34,
and ResNet-50, at sparsities from 90\% to 95.5\%. Our results show that
allocation choice systematically changes post-repair accuracy at the same
global sparsity, that a repair-sensitive transition regime exists in which BN
recalibration fails but ASR still recovers nontrivial accuracy, and that the
width and location of this regime vary with architecture and dataset difficulty.
Additional validations on ImageNet-100 and DenseNet-121 further characterize
how recoverability varies with data scale and connectivity pattern.
\end{abstract}

\section{Introduction}

Neural network pruning reduces inference cost by removing weights, but
aggressive pruning at high sparsity can cause severe accuracy collapse, a
regime in which the pruned model performs at or near random-chance levels,
well beyond ordinary accuracy degradation. Recovering from this collapse
typically requires fine-tuning with labeled data, which may be unavailable
when models are deployed in data-restricted environments or when the original
training pipeline is inaccessible.

Prior work on pruning has concentrated primarily on \emph{how to prune}.
Sparsity allocation strategies such as ERK~\citep{evci2020rigging} and
LAMP~\citep{lee2021lamp} determine how density budgets are distributed across
layers, and magnitude-based criteria remove weights of smallest absolute
value. These methods address the allocation problem but do not specify what
should be done \emph{after} pruning, particularly in the label-free setting.

Recent work on Adaptive Signal Resuscitation (ASR) fills this gap by
formulating label-free post-pruning recovery as a channelwise
activation-statistic repair problem~\citep{zhan2026asr}. ASR estimates a
variance-matching correction for each output channel and stabilizes it with a
data-driven shrinkage rule, requiring only forward passes on a small
calibration set. This repair operator has been shown to substantially
outperform BatchNorm (BN) recalibration at high sparsity across several
architectures and datasets.

\begin{figure}[t]
  \centering
  \includegraphics[width=\textwidth]{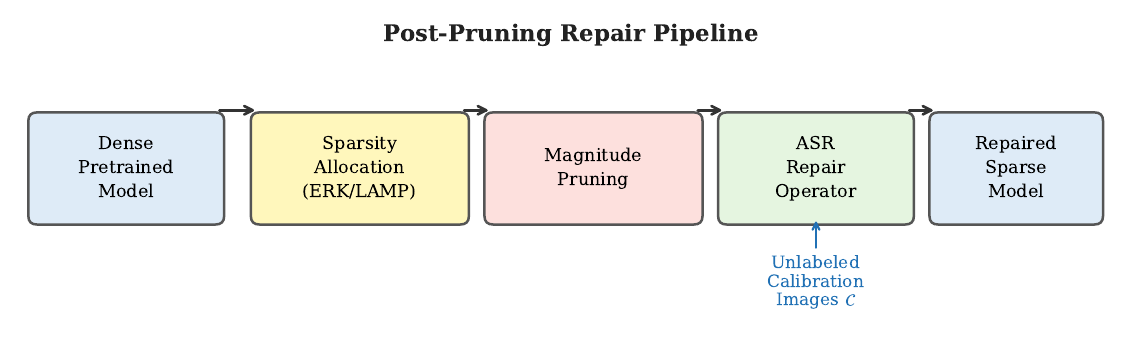}
  \caption{The post-pruning repair pipeline studied in this work. A dense
    pretrained model is pruned via a sparsity allocation rule (ERK or LAMP)
    followed by magnitude pruning. The ASR repair operator~\citep{zhan2026asr}
    is then applied as a fixed post-processing stage using only unlabeled
    calibration images $\calC$. We study how the allocation choice affects
    post-repair accuracy under this fixed backend.}
  \label{fig:pipeline}
\end{figure}

This paper asks a complementary question. Rather than asking \emph{how} to
repair a pruned model, we ask \emph{when} a pruned model remains repairable:
given a fixed label-free repair backend, how does the upstream sparsity
allocation affect the amount of activation signal that survives pruning and
can be recovered? We treat ASR as a fixed repair protocol and vary only the
allocation rule, examining how ERK and LAMP preserve or destroy the
channelwise activation signal that ASR can exploit. This perspective shifts
the focus from repair operator design to the recoverability consequences of
allocation choice. Figure~\ref{fig:pipeline} illustrates the overall pipeline.

\paragraph{Contributions.}
\begin{enumerate}[leftmargin=1.5em,itemsep=2pt]
  \item We study the interaction between layerwise sparsity allocation and
    label-free post-pruning repairability, using ASR~\citep{zhan2026asr} as
    a fixed repair backend rather than proposing a new repair operator.
  \item We show that allocation choice changes the amount of recoverable
    activation signal left after pruning: ERK and LAMP yield substantially
    different post-repair accuracies at the same global sparsity, with the
    direction of the advantage changing across architectures and datasets.
  \item We identify repair-sensitive transition regimes in which BN
    recalibration begins to fail but ASR still recovers nontrivial accuracy.
    Outside this band, either BN already performs well or the sparse model is
    close to irrecoverable collapse; the width and location of the transition
    vary with architecture, allocation, and dataset difficulty.
  \item Across ResNet-18, ResNet-34, and ResNet-50 on CIFAR-10, CIFAR-100,
    and Imagenette, with additional ImageNet-100 and DenseNet-121 validations,
    we show that post-repair recoverability varies systematically with
    allocation, sparsity, dataset difficulty, and architecture, and that the
    appropriate ASR variant (aggressive versus conservative) depends on the
    activation landscape left by the allocation.
\end{enumerate}

\section{Related Work}

\paragraph{Sparsity allocation.}
ERK~\citep{evci2020rigging} distributes the density budget across layers
using a structural rule based on layer shape. LAMP~\citep{lee2021lamp} derives
a layer-adaptive magnitude pruning score by accounting for layerwise
distortion. Both address the allocation problem but do not specify a
post-pruning repair operator. This work treats ERK and LAMP as upstream
allocation rules and examines how their structural differences affect the
activation signal available to a fixed downstream repair stage.

\paragraph{Post-pruning fine-tuning.}
The standard practice after pruning is to fine-tune the sparse model with
labeled data and gradient-based
optimization~\citep{han2015learning,frankle2019lottery}. This requires labeled
training data and a full backward pass. Our setting prohibits both, and we
treat training-based recovery as a distinct operational regime rather than a
direct baseline.

\paragraph{BatchNorm recalibration.}
Recomputing BN statistics from a calibration set without labels or gradients
is a common lightweight adjustment after pruning or
quantization~\citep{nagel2020adaround}. We treat BN recalibration as the
primary label-free baseline and characterize the sparsity regime in which it
begins to fail, which defines the lower boundary of the repair-sensitive
transition region.

\paragraph{Post-training quantization and calibration repair.}
\citet{nagel2019dfq} propose data-free quantization via weight equalization
and bias correction. \citet{lazarevich2021posttraining} study post-training
pruning through layerwise calibration, including bias correction followed by
BN recalibration. These approaches target mean-shift distortion; ASR differs
by addressing variance-level degradation, which we find necessary for
reliable recovery in high-sparsity regimes.

\paragraph{Label-free post-pruning repair.}
Adaptive Signal Resuscitation (ASR) proposes a channelwise repair operator
for sparse vision networks, showing that matching the granularity of repair to
the granularity of damage substantially improves recovery over layer-wise
methods~\citep{zhan2026asr}. The present work uses ASR as a fixed backend
and studies a different question: how does upstream allocation determine the
amount of signal available for ASR to repair?

\paragraph{Activation-statistic correction.}
Activation statistics have been used in related contexts to reduce
distributional mismatch after model transformation. REPAIR renormalizes
preactivations against reference statistics to mitigate variance distortion
in model merging~\citep{jordan2023repair}. ASR shares the calibration-only
philosophy but applies channelwise correction to pruning-induced activation
damage. The present work does not introduce a new correction operator; instead,
it uses ASR to study how allocation choices determine the activation landscape
available for repair.

\paragraph{Concurrent weight-rescaling approaches.}
Recent concurrent work proposes energy compensation for post-training LLM
pruning using column-wise and row-wise weight energy ratios to rescale pruned
weights~\citep{yu2026statistical}. That correction is weight-based and
data-free, applied at moderate sparsity on transformer architectures without
BatchNorm. The setting differs from ours along three axes: we use activation
statistics from calibration images, target extreme sparsity regimes where
activation collapse is severe, and operate on convolutional networks with
BatchNorm layers.

\section{Problem Setting}

Let $f_\theta$ denote a dense pretrained network with parameters $\theta$,
and let $f_{\hat\theta}$ denote the pruned sparse model obtained by applying
a sparsity allocation $\mathcal{A} \in \{\text{ERK}, \text{LAMP}\}$ followed
by magnitude pruning to a target global sparsity $s \in (0,1)$. Let
$\calC = \{x_1, \ldots, x_n\}$ be a small set of unlabeled calibration
images, with $n = 128$ in all experiments.

The label-free repair operator $\calR_{\text{ASR}}$~\citep{zhan2026asr}
maps $\hat\theta$ to repaired parameters using only $\calC$ and the dense
reference $f_\theta$, without labels or gradient computation. We fix
$\calR_{\text{ASR}}$ and study how the allocation $\mathcal{A}$ affects the
post-repair accuracy $\mathrm{acc}(f_{\calR_{\text{ASR}}(\hat\theta)})$ as a
function of sparsity $s$, architecture, and dataset. This formulation isolates
the allocation's contribution to recoverability from the repair operator
itself.

\section{Repair Backend: Activation-Statistic Repair}
\label{sec:method}

We use ASR as the fixed label-free repair backend throughout this work,
following \citet{zhan2026asr}. We summarize the repair operator below for
completeness, as all allocation comparisons are evaluated after the same
ASR-based repair stage.

\subsection{Motivation}

Pruning removes a subset of convolutional weights in each layer. For a given
output channel, the surviving weights may account for only a small fraction
of the original signal contribution, yielding activations with substantially
lower variance than those of the dense model. When this variance reduction is
severe, downstream BatchNorm layers and nonlinearities receive signals outside
the range induced by the dense model, contributing to prediction collapse.

\begin{figure}[t]
  \centering
  \includegraphics[width=\textwidth]{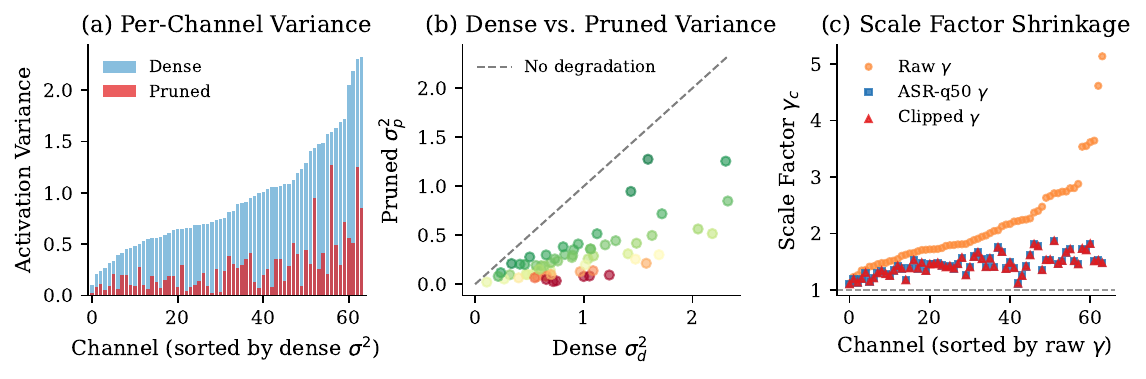}
  \caption{Per-channel variance degradation and scale factor shrinkage at
    92.5\% sparsity. \textbf{(a)}~Per-channel activation variance before and
    after pruning, sorted by dense variance. \textbf{(b)}~Scatter of dense
    versus pruned variance; channels below the diagonal have lost activation
    magnitude. \textbf{(c)}~Comparison of raw $\gamma^{\mathrm{raw}}_c$,
    shrinkage-adjusted $\gamma^{q50}_c$, and clipped
    $\gamma^{\mathrm{clip}}_c$ scale factors. The shrinkage prior suppresses
    large corrections in severely damaged channels; clipping further bounds
    the correction.}
  \label{fig:variance}
\end{figure}

The ratio of dense-to-pruned activation variance, computed on shared
calibration images, provides a channel-specific signal for how much the
pruned channel's activation scale has been degraded. Rescaling the surviving
output-channel weights to partially restore this variance can meaningfully
improve the sparse model's predictions without labels or gradient updates.
Figure~\ref{fig:variance} illustrates per-channel variance degradation and
the effect of the two shrinkage strategies on the scale factor $\gamma_c$.

\subsection{Channelwise Variance Estimation}

For each convolutional layer $l$ and output channel $c$, let $A^d_{l,c}$ and
$A^p_{l,c}$ denote the pre-BatchNorm activations produced by the dense and
pruned models on $\calC$. The per-channel variances are
\begin{equation}
  \sigma^2_{d,c} = \Var\!\left[A^d_{l,c}\right], \qquad
  \sigma^2_{p,c} = \Var\!\left[A^p_{l,c}\right].
  \label{eq:variances}
\end{equation}
The naive variance-matching scale factor for channel $c$ is
\begin{equation}
  \gamma^{\mathrm{raw}}_c =
    \sqrt{\frac{\sigma^2_{d,c}}{\sigma^2_{p,c} + \varepsilon}},
  \label{eq:gamma_raw}
\end{equation}
where $\varepsilon$ is a small constant for numerical stability. When
$\sigma^2_{p,c}$ is very small, as occurs in nearly collapsed channels, this
ratio can become large and may amplify noise rather than recover useful signal.

\subsection{ASR-q50: Median-Variance Shrinkage}

To attenuate noise amplification in severely damaged channels, ASR applies
a shrinkage prior based on the median pruned variance across channels:
\begin{equation}
  \lambda = \Med_c\!\left\{\sigma^2_{p,c}\right\}, \qquad
  \rho_c = \frac{\sigma^2_{p,c}}{\sigma^2_{p,c} + \lambda}.
  \label{eq:shrinkage_weight}
\end{equation}
The shrinkage-adjusted scale factor is then
\begin{equation}
  \gamma^{q50}_c = \rho_c \cdot \gamma^{\mathrm{raw}}_c + (1 - \rho_c).
  \label{eq:gamma_q50}
\end{equation}
The weight $\rho_c$ approaches 1 for channels whose pruned variance exceeds
the median and approaches 0 for the most severely damaged channels, smoothly
interpolating toward no correction at the extremes.

\subsection{Clipped ASR: Conservative Correction}

In severe collapse regimes, ASR-q50 may still over-correct certain channels.
Clipped ASR constrains the scale factor by
\begin{equation}
  \gamma^{\mathrm{clip}}_c =
    \clamp\!\left(\gamma^{q50}_c,\; 0.5,\; 2.0\right).
  \label{eq:gamma_clip}
\end{equation}
The bounds $[0.5, 2.0]$ prevent any channel from being scaled by more than a
factor of two in either direction, trading peak recovery magnitude for
stability. The two variants represent complementary points in the
aggressiveness-stability tradeoff; comparing their relative performance under
different allocations is one way to characterize the activation landscape left
by the allocation.

\subsection{Weight Rescaling and BN Recalibration}
\label{sec:weight_rescaling}

After computing $\gamma_c$, the surviving weights of output channel $c$ are
rescaled and the bias term is adjusted to align the repaired activation mean
with the dense activation mean:
\begin{align}
  \hat{w}_{l,c} &\leftarrow \gamma_c\, w_{l,c},
  \label{eq:weight_rescale}\\
  \hat{b}_{l,c} &\leftarrow \gamma_c\, b_{l,c}
    + \mu^d_{l,c} - \gamma_c\, \mu^p_{l,c},
  \label{eq:bias_adjust}
\end{align}
where $\mu^d_{l,c}$ and $\mu^p_{l,c}$ are the dense and pruned activation
means for channel $c$ on $\calC$. If the original convolution has no bias
term, we set $b_{l,c} = 0$ and insert the repaired bias. A BN recalibration
pass over 20 mini-batches of unlabeled images then updates each BatchNorm
layer's running statistics. The complete procedure requires only forward
passes and involves no gradient computation.

\section{Experimental Setup}

\paragraph{Architectures and datasets.}
Main experiments are conducted on ResNet-18, ResNet-34, and ResNet-50 across
CIFAR-10, CIFAR-100, and Imagenette. Two additional validations extend the
scope: an ImageNet-100 experiment with ResNet-18, and a DenseNet-121
experiment on CIFAR-10 and CIFAR-100 as a non-residual architecture check.

\paragraph{Sparsity allocations.}
Two standard allocations are compared: ERK~\citep{evci2020rigging} and
LAMP~\citep{lee2021lamp}. Unstructured magnitude pruning is applied at the
target global sparsity. Both allocations are evaluated under the same ASR
repair protocol so that any differences in post-repair accuracy are
attributable to the allocation rather than the repair operator.

\paragraph{Target sparsities.}
For ResNet-50, CIFAR-10 and Imagenette are evaluated at 90.0\%, 92.5\%, and
95.0\% sparsity, while CIFAR-100 is evaluated at 92.5\%, 95.0\%, and 95.5\%.
For ResNet-18 and ResNet-34, all three datasets are evaluated at 90.0\%,
92.5\%, and 95.0\%. The ImageNet-100 validation uses the same three levels.
The DenseNet-121 evaluation on CIFAR-10 uses 90.0\%, 92.5\%, and 95.0\%; the
CIFAR-100 evaluation uses 70.0\%, 80.0\%, 85.0\%, and 90.0\% to capture the
transition into collapse, which occurs at lower sparsity for this architecture.

\paragraph{Calibration.}
All repair methods use $n = 128$ unlabeled images drawn from the training
set. BN recalibration uses 20 mini-batches. No labels and no gradient updates
are used at any point during repair.

\paragraph{Baselines.}
Four label-free repair strategies are compared alongside the two ASR variants:
(i)~\emph{No repair}, direct evaluation of the pruned model;
(ii)~\emph{BN only}, BN recalibration with no weight modification;
(iii)~\emph{Bias correction + BN}, per-channel mean-shift correction followed
by BN recalibration; and (iv)~\emph{Affine calibration + BN}, per-channel
affine output calibration followed by BN recalibration.

\section{Results}

Aggregating over the 54 main ResNet allocation-sparsity settings, the best
ASR variant improves over BN-only repair in every condition. The mean gain
over BN recalibration is 17.4 percentage points, with a minimum gain of 1.3
points. ASR also outperforms the strongest simple calibration baseline in all
54 settings, with an average margin of 13.6 points. These aggregate results
establish that ASR provides a stable repair backend across the main allocation
grid, making it meaningful to analyze how upstream allocation changes
post-repair recoverability.

\subsection{Repair Comparison Under Fixed Allocation}
\label{sec:repair_comparison}

\begin{figure}[t]
  \centering
  \includegraphics[width=\textwidth]{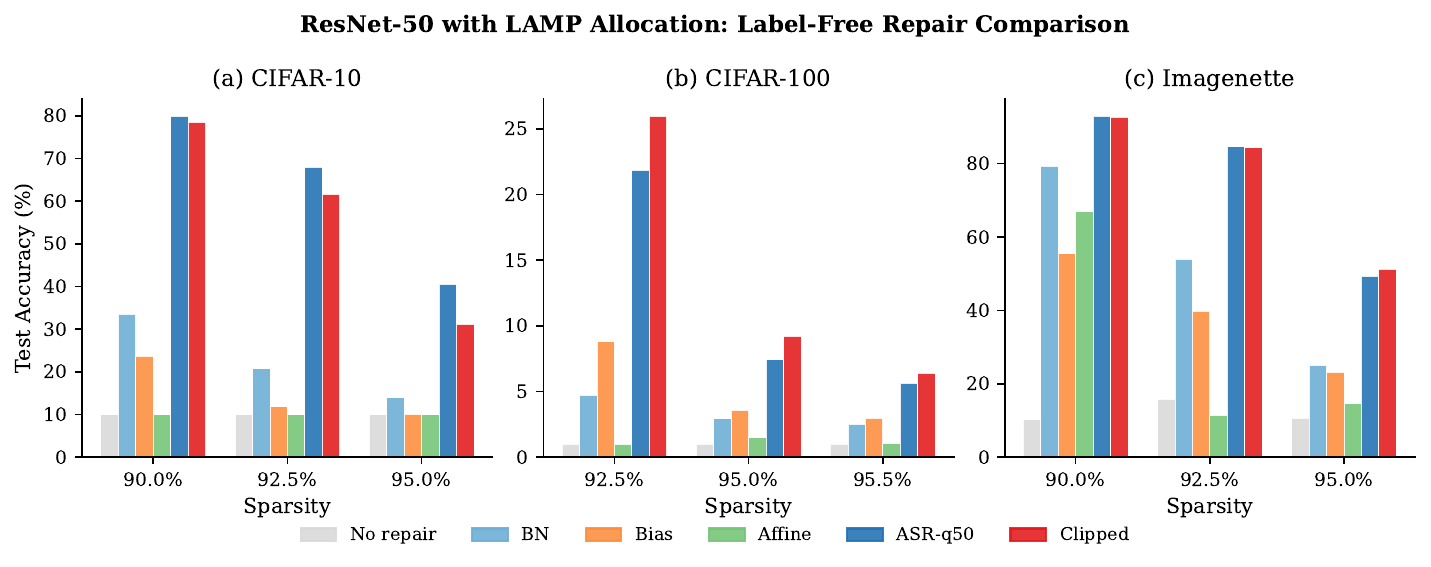}
  \caption{Repair comparison on ResNet-50 with LAMP allocation across
    CIFAR-10, CIFAR-100, and Imagenette. ASR-q50 and Clipped ASR consistently
    outperform all four baselines across sparsity levels. The large gap over
    BN-only recalibration marks the repair-sensitive regime.}
  \label{fig:main_results}
\end{figure}

Table~\ref{tab:repair_comparison} compares label-free repair methods under
both ERK and LAMP allocations across all three ResNet architectures, isolating
the repair operator by holding the upstream pruning allocation fixed.
Figure~\ref{fig:main_results} summarizes the ResNet-50 results visually.

\paragraph{CIFAR-10.}
On ResNet-50 at 92.5\% sparsity, BN recalibration reaches 20.72\%, while
ASR-q50 and Clipped ASR reach 67.96\% and 61.59\%, gains of approximately 47
and 41 percentage points over BN only. Bias correction (11.79\%) and affine
calibration (10.00\%, near random chance) both fail substantially. ResNet-34
presents a notably different pattern: affine calibration reaches 87.24\% at
92.5\%, outperforming the BN-only baseline of 62.07\%, yet ASR-q50 still
leads at 89.27\%. On ResNet-18, BN only reaches 33.95\% at 92.5\%, while
ASR-q50 reaches 66.79\% and Clipped ASR reaches 63.33\%. ASR-q50 is
generally stronger than Clipped ASR on CIFAR-10, consistent with the
interpretation that the allocation leaves enough recoverable signal for
aggressive correction to be stable.

\paragraph{CIFAR-100.}
On ResNet-50 at 92.5\% sparsity, the pruned model achieves 1.00\% without
repair. BN recalibration raises this to only 4.70\%, marking the entry into
the repair-sensitive regime; ASR-q50 reaches 21.85\% and Clipped ASR reaches
25.96\%. Affine calibration nearly universally fails. ResNet-34 shows notably
stronger recovery across all methods; at 90.0\% under LAMP, BN only reaches
44.96\% and ASR-q50 reaches 59.81\%. On ResNet-18 under LAMP at 92.5\%, BN
only reaches 18.45\%, ASR-q50 reaches 26.48\%, and Clipped ASR reaches
26.31\%. On ResNet-18 and ResNet-50, Clipped ASR is competitive with or
stronger than ASR-q50 on CIFAR-100, suggesting that the allocation leaves a
more fragile activation landscape on these architectures at high sparsity.

\paragraph{Imagenette.}
Both ASR variants produce strong recovery across architectures. On ResNet-50
at 92.5\% under LAMP, BN only reaches 53.96\%, while ASR-q50 and Clipped ASR
reach 84.64\% and 84.31\%. On ResNet-18, BN only already reaches 83.82\%, and
ASR-q50 and Clipped ASR improve further to 88.84\% and 89.20\%, confirming
that ASR provides gains even when the BN baseline is relatively strong.
ResNet-34 at 92.5\% under LAMP shows ASR-q50 and Clipped ASR reaching 90.04\%
and 90.34\%, compared to 84.23\% for BN only.

\begin{table*}[!t]
\caption{Repair comparison under fixed ERK and LAMP allocations across ResNet
  depths. Best result in each row is \textbf{bold}.}
\label{tab:repair_comparison}
\begin{center}
\scriptsize
\setlength{\tabcolsep}{2.3pt}
\begin{adjustbox}{max width=\textwidth}
\begin{tabular}{llllcccccc}
\toprule
Arch & Dataset & Sparsity & Alloc.
  & No repair & BN & Bias & Affine & ASR-q50 & Clip \\
\midrule
\multirow{18}{*}{ResNet-18}
& \multirow{6}{*}{CIFAR-10}
  & 90.0\% & ERK  & 10.00 & 59.39 & 42.76 & 17.99 & \textbf{76.92} & 75.39 \\
& & 90.0\% & LAMP & 10.44 & 63.87 & 62.16 & 53.16 & \textbf{79.91} & 78.55 \\
& & 92.5\% & ERK  & 10.00 & 32.78 & 18.54 & 10.00 & \textbf{61.97} & 59.02 \\
& & 92.5\% & LAMP & 10.00 & 33.95 & 46.13 & 17.89 & \textbf{66.79} & 63.33 \\
& & 95.0\% & ERK  & 10.00 & 14.73 & 13.04 & 10.00 & \textbf{34.04} & 31.99 \\
& & 95.0\% & LAMP & 10.00 & 13.59 & 16.45 & 10.00 & \textbf{41.05} & 36.20 \\
\cmidrule(lr){2-10}
& \multirow{6}{*}{CIFAR-100}
  & 90.0\% & ERK  & 1.14 & 31.82 &  9.08 &  1.90 & \textbf{38.13} & 38.11 \\
& & 90.0\% & LAMP & 1.44 & 33.68 & 12.64 &  5.92 & 40.09 & \textbf{40.25} \\
& & 92.5\% & ERK  & 1.11 & 17.26 &  2.69 &  1.12 & \textbf{24.34} & 24.19 \\
& & 92.5\% & LAMP & 1.17 & 18.45 &  6.71 &  1.18 & \textbf{26.48} & 26.31 \\
& & 95.0\% & ERK  & 0.95 &  7.06 &  1.44 &  1.00 &  9.40 & \textbf{9.45} \\
& & 95.0\% & LAMP & 1.06 &  5.72 &  2.55 &  1.00 & \textbf{10.96} & 10.89 \\
\cmidrule(lr){2-10}
& \multirow{6}{*}{Imagenette}
  & 90.0\% & ERK  & 18.90 & 90.04 & 74.45 & 44.23 & 91.87 & \textbf{92.00} \\
& & 90.0\% & LAMP & 20.82 & 91.69 & 79.52 & 82.37 & 92.94 & \textbf{93.15} \\
& & 92.5\% & ERK  & 13.68 & 82.09 & 36.82 & 11.90 & 86.01 & \textbf{86.24} \\
& & 92.5\% & LAMP & 17.50 & 83.82 & 46.47 & 17.48 & 88.84 & \textbf{89.20} \\
& & 95.0\% & ERK  & 10.47 & 58.27 & 16.61 &  9.10 & 68.51 & \textbf{68.66} \\
& & 95.0\% & LAMP & 10.19 & 60.69 & 11.92 &  9.10 & 73.10 & \textbf{73.20} \\
\midrule
\multirow{18}{*}{ResNet-34}
& \multirow{6}{*}{CIFAR-10}
  & 90.0\% & ERK  & 22.21 & 84.50 & 70.22 & 88.79 & \textbf{90.12} & 89.87 \\
& & 90.0\% & LAMP & 15.34 & 80.34 & 70.29 & 89.53 & \textbf{90.43} & 89.89 \\
& & 92.5\% & ERK  & 17.09 & 72.02 & 47.57 & 85.63 & \textbf{89.05} & 88.15 \\
& & 92.5\% & LAMP & 11.65 & 62.07 & 50.96 & 87.24 & \textbf{89.27} & 87.12 \\
& & 95.0\% & ERK  & 12.46 & 39.16 & 29.10 & 72.26 & \textbf{85.47} & 81.57 \\
& & 95.0\% & LAMP & 10.34 & 28.93 & 35.38 & 75.57 & \textbf{85.81} & 77.09 \\
\cmidrule(lr){2-10}
& \multirow{6}{*}{CIFAR-100}
  & 90.0\% & ERK  &  3.73 & 49.77 & 23.87 & 55.68 & \textbf{61.38} & 61.02 \\
& & 90.0\% & LAMP &  3.59 & 44.96 & 20.16 & 55.62 & \textbf{59.81} & 58.88 \\
& & 92.5\% & ERK  &  2.02 & 33.32 & 10.87 & 46.18 & \textbf{55.25} & 54.30 \\
& & 92.5\% & LAMP &  2.28 & 27.58 &  8.66 & 42.68 & \textbf{51.83} & 49.64 \\
& & 95.0\% & ERK  &  1.85 & 13.17 &  4.02 & 26.30 & \textbf{41.20} & 40.34 \\
& & 95.0\% & LAMP &  1.95 & 10.93 &  4.53 & 24.66 & \textbf{38.07} & 35.50 \\
\cmidrule(lr){2-10}
& \multirow{6}{*}{Imagenette}
  & 90.0\% & ERK  & 31.67 & 89.73 & 50.83 & 87.52 & \textbf{91.06} & 91.03 \\
& & 90.0\% & LAMP & 39.26 & 89.20 & 39.39 & 89.53 & 90.83 & \textbf{90.96} \\
& & 92.5\% & ERK  & 13.86 & 86.45 & 19.77 & 80.20 & 90.09 & \textbf{90.19} \\
& & 92.5\% & LAMP & 27.01 & 84.23 & 14.06 & 85.81 & 90.04 & \textbf{90.34} \\
& & 95.0\% & ERK  & 10.55 & 73.43 & 10.34 & 55.87 & 88.00 & \textbf{88.28} \\
& & 95.0\% & LAMP & 16.25 & 66.78 & 10.11 & 66.93 & 87.80 & \textbf{88.15} \\
\midrule
\multirow{18}{*}{ResNet-50}
& \multirow{6}{*}{CIFAR-10}
  & 90.0\% & ERK  & 10.00 & 28.19 & 10.58 & 10.00 & \textbf{70.24} & 64.81 \\
& & 90.0\% & LAMP & 10.00 & 33.44 & 23.63 & 10.00 & \textbf{79.92} & 78.51 \\
& & 92.5\% & ERK  & 10.00 & 19.02 & 12.82 & 10.00 & \textbf{54.56} & 47.85 \\
& & 92.5\% & LAMP & 10.00 & 20.72 & 11.79 & 10.00 & \textbf{67.96} & 61.59 \\
& & 95.0\% & ERK  & 10.00 & 15.08 &  9.38 & 10.00 & \textbf{33.87} & 31.62 \\
& & 95.0\% & LAMP & 10.00 & 13.95 & 10.09 & 10.00 & \textbf{40.55} & 31.10 \\
\cmidrule(lr){2-10}
& \multirow{6}{*}{CIFAR-100}
  & 92.5\% & ERK  & 1.63 &  5.70 &  4.63 &  1.00 & 18.88 & \textbf{20.69} \\
& & 92.5\% & LAMP & 1.00 &  4.70 &  8.79 &  1.00 & 21.85 & \textbf{25.96} \\
& & 95.0\% & ERK  & 1.00 &  2.48 &  1.55 &  1.00 &  7.13 & \textbf{8.68} \\
& & 95.0\% & LAMP & 1.00 &  2.93 &  3.58 &  1.49 &  7.43 & \textbf{9.20} \\
& & 95.5\% & ERK  & 1.00 &  2.12 &  1.60 &  1.23 &  5.45 & \textbf{7.04} \\
& & 95.5\% & LAMP & 1.00 &  2.51 &  2.98 &  1.06 &  5.65 & \textbf{6.40} \\
\cmidrule(lr){2-10}
& \multirow{6}{*}{Imagenette}
  & 90.0\% & ERK  & 13.71 & 65.35 & 26.96 & 17.43 & 88.38 & \textbf{89.10} \\
& & 90.0\% & LAMP & 10.24 & 79.34 & 55.47 & 66.85 & \textbf{92.97} & 92.64 \\
& & 92.5\% & ERK  & 10.68 & 44.51 & 16.79 & 19.80 & 75.52 & \textbf{75.87} \\
& & 92.5\% & LAMP & 15.80 & 53.96 & 39.85 & 11.39 & \textbf{84.64} & 84.31 \\
& & 95.0\% & ERK  & 10.68 & 16.64 & 15.80 &  9.91 & \textbf{44.33} & 44.05 \\
& & 95.0\% & LAMP & 10.68 & 24.99 & 23.16 & 14.60 & 49.25 & \textbf{51.24} \\
\bottomrule
\end{tabular}
\end{adjustbox}
\end{center}
\end{table*}

\subsection{Allocation-Repair Interaction}
\label{sec:pipeline_comparison}

\begin{figure}[t]
  \centering
  \includegraphics[width=\textwidth]{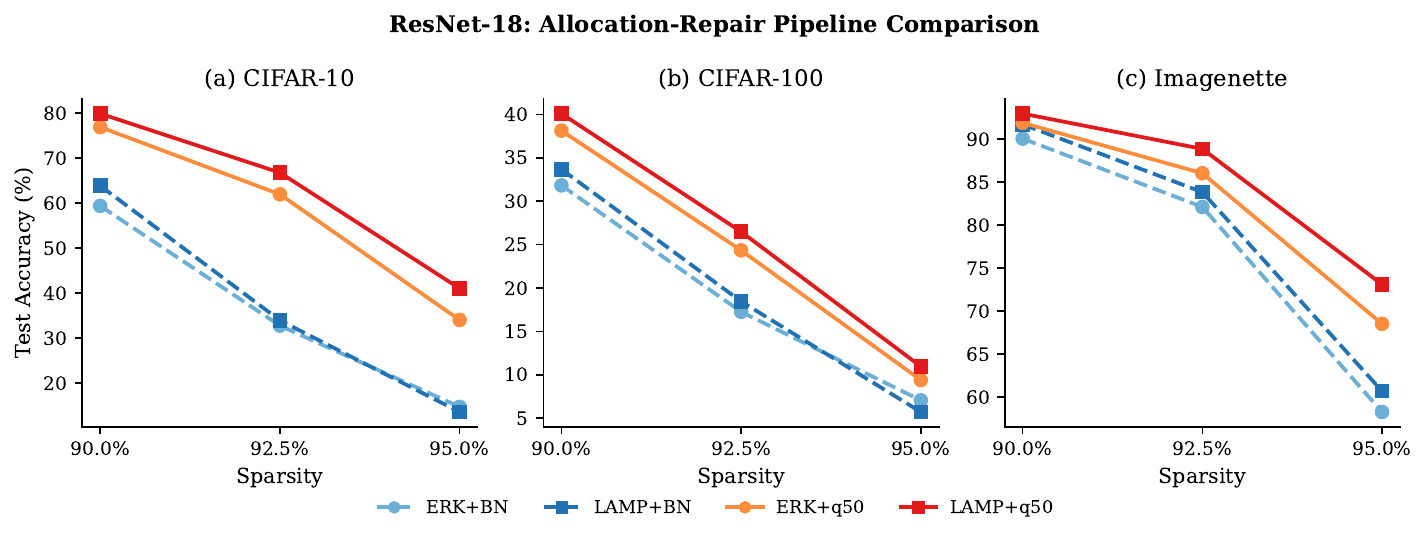}
  \caption{Allocation-repair interaction on ResNet-18 across CIFAR-10,
    CIFAR-100, and Imagenette. Solid lines correspond to ASR-q50-repaired
    models; dashed lines correspond to BN-only baselines. The gap between
    ERK and LAMP post-repair accuracy varies across datasets and sparsity
    levels, demonstrating that allocation choice shapes recoverability.}
  \label{fig:pipeline_comparison}
\end{figure}

Table~\ref{tab:pipeline_comparison} directly compares ERK and LAMP under the
same ASR repair backend, making allocation the only variable. This is the
central comparison for studying how allocation affects recoverability.
Figure~\ref{fig:pipeline_comparison} presents the ResNet-18 results as line
plots across sparsity levels.

\paragraph{CIFAR-10.}
On ResNet-50, LAMP consistently yields higher post-repair accuracy than ERK
at the same sparsity: at 90.0\%, LAMP + ASR-q50 reaches 79.92\% versus
70.24\% for ERK + ASR-q50, a gap of nearly 10 points. ResNet-34 tells a
different story: LAMP + ASR-q50 and ERK + ASR-q50 are nearly tied at 90.0\%
(90.43\% versus 90.12\%), but ERK begins to pull ahead at 95.0\% (85.47\%
versus 85.81\%), suggesting that ERK's more uniform layer-level density
preserves more repairable signal at the highest sparsities on this
architecture. On ResNet-18, LAMP is also slightly stronger at all three levels.

\paragraph{CIFAR-100.}
The allocation dependence is sharpest on CIFAR-100, the most challenging
dataset. On ResNet-50 at 92.5\%, LAMP + Clipped ASR reaches 25.96\% versus
20.69\% for ERK + Clipped ASR, a 5.3-point gap from allocation alone. The
direction reverses on ResNet-34: ERK + ASR-q50 leads at every sparsity level
(61.38\% versus 59.81\% at 90.0\%, widening to 41.20\% versus 38.07\% at
95.0\%). This reversal indicates that which allocation preserves more
recoverable signal depends on architecture, and cannot be determined from
global sparsity alone.

\paragraph{Imagenette}
LAMP allocations generally preserve more repairable signal than ERK on both
ResNet-18 and ResNet-50. On ResNet-18 at 95.0\%, LAMP + ASR-q50 reaches
73.10\% versus 68.51\% for ERK + ASR-q50. The gap is smaller on Imagenette
than on CIFAR-100, consistent with the interpretation that easier tasks leave
more residual signal regardless of allocation, reducing allocation sensitivity.

\begin{table*}[t]
\caption{Allocation-repair interaction: ERK versus LAMP under a fixed ASR
  repair backend across ResNet depths. Best result in each row is \textbf{bold}.}
\label{tab:pipeline_comparison}
\begin{center}
\scriptsize
\setlength{\tabcolsep}{2.5pt}
\begin{adjustbox}{max width=\textwidth}
\begin{tabular}{lllcccccc}
\toprule
Arch & Dataset & Sparsity
  & ERK+BN & ERK+q50 & ERK+Clip
  & LAMP+BN & LAMP+q50 & LAMP+Clip \\
\midrule
\multirow{9}{*}{ResNet-18}
& \multirow{3}{*}{CIFAR-10}
  & 90.0\% & 59.39 & 76.92 & 75.39 & 63.87 & \textbf{79.91} & 78.55 \\
& & 92.5\% & 32.78 & 61.97 & 59.02 & 33.95 & \textbf{66.79} & 63.33 \\
& & 95.0\% & 14.73 & 34.04 & 31.99 & 13.59 & \textbf{41.05} & 36.20 \\
\cmidrule(lr){2-9}
& \multirow{3}{*}{CIFAR-100}
  & 90.0\% & 31.82 & 38.13 & 38.11 & 33.68 & 40.09 & \textbf{40.25} \\
& & 92.5\% & 17.26 & 24.34 & 24.19 & 18.45 & \textbf{26.48} & 26.31 \\
& & 95.0\% &  7.06 &  9.40 &  9.45 &  5.72 & \textbf{10.96} & 10.89 \\
\cmidrule(lr){2-9}
& \multirow{3}{*}{Imagenette}
  & 90.0\% & 90.04 & 91.87 & 92.00 & 91.69 & 92.94 & \textbf{93.15} \\
& & 92.5\% & 82.09 & 86.01 & 86.24 & 83.82 & 88.84 & \textbf{89.20} \\
& & 95.0\% & 58.27 & 68.51 & 68.66 & 60.69 & 73.10 & \textbf{73.20} \\
\midrule
\multirow{9}{*}{ResNet-34}
& \multirow{3}{*}{CIFAR-10}
  & 90.0\% & 84.50 & 90.12 & 89.87 & 80.34 & \textbf{90.43} & 89.89 \\
& & 92.5\% & 72.02 & 89.05 & 88.15 & 62.07 & \textbf{89.27} & 87.12 \\
& & 95.0\% & 39.16 & 85.47 & 81.57 & 28.93 & \textbf{85.81} & 77.09 \\
\cmidrule(lr){2-9}
& \multirow{3}{*}{CIFAR-100}
  & 90.0\% & 49.77 & \textbf{61.38} & 61.02 & 44.96 & 59.81 & 58.88 \\
& & 92.5\% & 33.32 & \textbf{55.25} & 54.30 & 27.58 & 51.83 & 49.64 \\
& & 95.0\% & 13.17 & \textbf{41.20} & 40.34 & 10.93 & 38.07 & 35.50 \\
\cmidrule(lr){2-9}
& \multirow{3}{*}{Imagenette}
  & 90.0\% & 89.73 & \textbf{91.06} & 91.03 & 89.20 & 90.83 & 90.96 \\
& & 92.5\% & 86.45 & 90.09 & 90.19 & 84.23 & 90.04 & \textbf{90.34} \\
& & 95.0\% & 73.43 & 88.00 & \textbf{88.28} & 66.78 & 87.80 & 88.15 \\
\midrule
\multirow{9}{*}{ResNet-50}
& \multirow{3}{*}{CIFAR-10}
  & 90.0\% & 28.19 & 70.24 & 64.81 & 33.44 & \textbf{79.92} & 78.51 \\
& & 92.5\% & 19.02 & 54.56 & 47.85 & 20.72 & \textbf{67.96} & 61.59 \\
& & 95.0\% & 15.08 & 33.87 & 31.62 & 13.95 & \textbf{40.55} & 31.10 \\
\cmidrule(lr){2-9}
& \multirow{3}{*}{CIFAR-100}
  & 92.5\% &  5.70 & 18.88 & 20.69 &  4.70 & 21.85 & \textbf{25.96} \\
& & 95.0\% &  2.48 &  7.13 &  8.68 &  2.93 &  7.43 & \textbf{9.20} \\
& & 95.5\% &  2.12 &  5.45 & \textbf{7.04} &  2.51 &  5.65 &  6.40 \\
\cmidrule(lr){2-9}
& \multirow{3}{*}{Imagenette}
  & 90.0\% & 65.35 & 88.38 & 89.10 & 79.34 & \textbf{92.97} & 92.64 \\
& & 92.5\% & 44.51 & 75.52 & 75.87 & 53.96 & \textbf{84.64} & 84.31 \\
& & 95.0\% & 16.64 & 44.33 & 44.05 & 24.99 & 49.25 & \textbf{51.24} \\
\bottomrule
\end{tabular}
\end{adjustbox}
\end{center}
\end{table*}

\subsection{ImageNet-100 Validation}
\label{sec:imagenet100}

Table~\ref{tab:imagenet100} reports results on ImageNet-100 with ResNet-18.
Both ASR variants consistently lead across all six conditions. The margins
over BN only are smaller than on CIFAR-100 at comparable sparsity, consistent
with the interpretation that ImageNet-scale models retain more distributed
signal per channel, leaving less room for variance-based correction to act and
narrowing the repair-sensitive transition region. The ERK-LAMP gap is also
smaller than on CIFAR-100, suggesting that allocation sensitivity contracts
when more residual signal survives pruning.

\begin{table}[t]
\caption{ImageNet-100 validation on ResNet-18. Best result in each row is \textbf{bold}.}
\label{tab:imagenet100}
\begin{center}
\scriptsize
\setlength{\tabcolsep}{4pt}
\begin{adjustbox}{max width=\linewidth}
\begin{tabular}{llcccccc}
\toprule
Sparsity & Alloc. & No repair & BN & Bias & Affine & ASR-q50 & Clip \\
\midrule
90.0\% & ERK  & 2.02 & 48.96 & 41.61 &  7.91 & 49.01 & \textbf{49.48} \\
90.0\% & LAMP & 2.78 & 49.75 & 43.83 & 24.81 & 49.75 & \textbf{50.17} \\
92.5\% & ERK  & 1.63 & 31.59 & 15.90 &  0.99 & 32.75 & \textbf{33.00} \\
92.5\% & LAMP & 1.97 & 30.99 & 21.94 &  2.17 & 32.13 & \textbf{32.71} \\
95.0\% & ERK  & 1.00 & 13.15 &  1.87 &  0.93 & 13.88 & \textbf{14.01} \\
95.0\% & LAMP & 1.15 & 11.95 &  5.21 &  1.26 & 13.10 & \textbf{13.20} \\
\bottomrule
\end{tabular}
\end{adjustbox}
\end{center}
\end{table}

\subsection{DenseNet-121 Validation}
\label{sec:densenet121}

Tables~\ref{tab:densenet_cifar10} and~\ref{tab:densenet_cifar100} report
results on DenseNet-121, probing whether the allocation-repair interaction
observed on ResNet architectures extends to non-residual connectivity. On
CIFAR-10, Clipped ASR provides substantial recovery at all three sparsity
levels and leads ASR-q50 throughout, reversing the ResNet-18 pattern and
indicating that DenseNet-121's dense connectivity leaves an activation
landscape where aggressive correction overshoots. This architecture-dependent
variant preference is itself a consequence of how allocation interacts with
connectivity structure.

On CIFAR-100, the repair-sensitive transition regime occurs at substantially
lower sparsity than in the ResNet experiments. At 70.0\% under LAMP, BN only
reaches 37.76\% and Clipped ASR reaches 49.76\%; the regime contracts sharply
by 80.0\%, where BN only manages 3.81\% under LAMP while Clipped ASR still
recovers 31.82\%. At 85.0\% and 90.0\%, the transition has passed and all
methods except Clipped ASR fall to near-random. This earlier onset of
irrecoverable collapse demonstrates that the allocation-repair transition
boundary is architecture-dependent and cannot be predicted from ResNet results
alone.

\begin{table}[t]
\caption{DenseNet-121 validation on CIFAR-10. Best result in each row is \textbf{bold}.}
\label{tab:densenet_cifar10}
\begin{center}
\scriptsize
\setlength{\tabcolsep}{4pt}
\begin{adjustbox}{max width=\linewidth}
\begin{tabular}{llcccccc}
\toprule
Sparsity & Alloc. & No repair & BN & Bias & Affine & ASR-q50 & Clip \\
\midrule
90.0\% & ERK  & 10.00 & 10.99 & 10.00 & 10.26 & 17.43 & \textbf{49.48} \\
90.0\% & LAMP & 10.00 & 10.21 & 10.00 & 11.89 & 15.02 & \textbf{28.82} \\
92.5\% & ERK  & 10.00 & 11.00 & 10.00 & 11.51 & 15.23 & \textbf{40.36} \\
92.5\% & LAMP & 10.00 & 10.00 & 10.00 & 10.45 & 17.75 & \textbf{31.87} \\
95.0\% & ERK  & 10.00 & 11.78 & 10.00 & 10.25 & 10.03 & \textbf{38.07} \\
95.0\% & LAMP & 10.00 & 10.00 & 10.00 & 14.08 & 10.00 & \textbf{32.02} \\
\bottomrule
\end{tabular}
\end{adjustbox}
\end{center}
\end{table}

\begin{table}[t]
\caption{DenseNet-121 validation on CIFAR-100. Best result in each row is \textbf{bold}.}
\label{tab:densenet_cifar100}
\begin{center}
\scriptsize
\setlength{\tabcolsep}{4pt}
\begin{adjustbox}{max width=\linewidth}
\begin{tabular}{llcccccc}
\toprule
Sparsity & Alloc. & No repair & BN & Bias & Affine & ASR-q50 & Clip \\
\midrule
70.0\% & ERK  &  3.27 & 26.98 &  7.46 & 10.11 & 34.64 & \textbf{36.57} \\
70.0\% & LAMP & 11.17 & 37.76 & 11.93 & 31.74 & 49.63 & \textbf{49.76} \\
80.0\% & ERK  &  1.02 &  2.68 &  2.49 &  3.38 &  6.73 & \textbf{13.10} \\
80.0\% & LAMP &  1.01 &  3.81 &  1.03 & 10.88 & 24.66 & \textbf{31.82} \\
85.0\% & ERK  &  1.00 &  1.13 &  1.58 &  2.62 &  1.16 &  \textbf{5.86} \\
85.0\% & LAMP &  1.00 &  1.04 &  1.00 &  3.60 &  4.45 & \textbf{14.72} \\
90.0\% & ERK  &  1.00 &  1.00 &  1.00 &  1.00 &  1.00 &  \textbf{2.85} \\
90.0\% & LAMP &  1.00 &  1.00 &  1.00 &  2.31 &  1.84 &  \textbf{4.20} \\
\bottomrule
\end{tabular}
\end{adjustbox}
\end{center}
\end{table}

\section{Discussion}

\paragraph{Allocation choice shapes the activation landscape available for repair.}
The ERK versus LAMP comparison under a fixed ASR backend demonstrates that
allocation is not allocation-agnostic from the repair perspective. LAMP
generally preserves more repairable signal on ResNet-18 and ResNet-50, but
ERK is consistently stronger on ResNet-34 / CIFAR-100. This architecture-
and dataset-dependent reversal suggests that the structural properties of an
allocation (how it distributes density across layer sizes and types) interact
with the architecture's connectivity in ways that determine how much
channelwise variance survives at any given global sparsity.

\paragraph{The repair-sensitive transition regime.}
Across all conditions, a consistent pattern emerges: a sparsity range in
which BN recalibration begins to fail but ASR still recovers nontrivial
accuracy. On ResNet-50 / CIFAR-100, this transition occurs around 92--93\%
sparsity; on DenseNet-121 / CIFAR-100, it occurs around 70--80\%. Within the
transition, the choice of allocation matters most, because marginal differences
in preserved channelwise variance translate into large differences in
post-repair accuracy. Outside the transition (either below it, where BN
already works well, or above it, where all methods collapse), allocation
sensitivity contracts.

\paragraph{The appropriate ASR variant depends on the allocation.}
ASR-q50 leads on CIFAR-10 across ResNet architectures, where the allocation
leaves enough recoverable signal for aggressive correction to be stable.
Clipped ASR leads on CIFAR-100 at high sparsity and on DenseNet-121
throughout, where the activation landscape after allocation is more fragile.
The relative ordering of the two variants therefore provides indirect evidence
about the character of the activation landscape that the allocation produces.

\paragraph{ASR is not a substitute for labeled fine-tuning.}
ASR occupies a distinct operational regime. Masked fine-tuning, which retrains
the sparse model with labeled data and gradient updates while preserving the
pruning mask, substantially outperforms ASR when it is available. At 92.5\%
sparsity on CIFAR-100 under LAMP, Clipped ASR reaches 25.96\%, while one
epoch of masked fine-tuning reaches 71.49\%. ASR is appropriate when labels
are unavailable and computational cost must be kept low.

\begin{figure}[t]
  \centering
  \includegraphics[width=\textwidth]{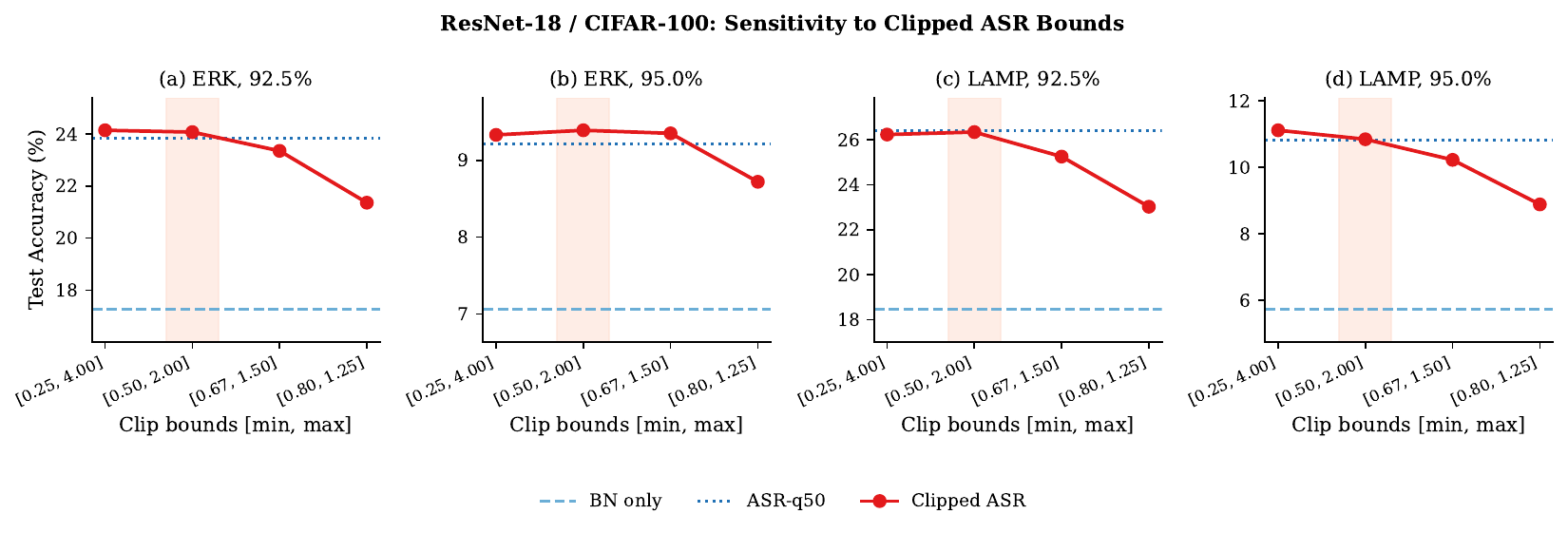}
  \caption{Sensitivity of Clipped ASR to clip bounds on ResNet-18 /
    CIFAR-100 under ERK and LAMP at 92.5\% and 95.0\% sparsity. The shaded
    column marks the default $[0.50, 2.00]$ setting. Performance is stable
    across the two widest configurations and degrades only when bounds are
    tightened substantially.}
  \label{fig:sensitivity}
\end{figure}

\paragraph{Sensitivity to clip bounds.}
Figure~\ref{fig:sensitivity} reports test accuracy on ResNet-18 / CIFAR-100
under four clip bound configurations: $[0.25, 4.00]$, $[0.50, 2.00]$,
$[0.67, 1.50]$, and $[0.80, 1.25]$. The two wider settings perform similarly
and both improve over ASR-q50 in several conditions, while tighter bounds
progressively reduce accuracy, most visibly under LAMP at 92.5\% sparsity
where $[0.80, 1.25]$ drops 3.4 percentage points below ASR-q50. The default
$[0.50, 2.00]$ bound sits in a flat region of this sensitivity curve,
suggesting that the exact value is not critical provided it is not set too
conservatively.

\section{Limitations and Future Work}

The variance-ratio mechanism assumes that the
dense model's activation statistics serve as an appropriate repair target, an
assumption that may not hold when the calibration distribution differs
substantially from the test distribution. The DenseNet-121 results indicate
that the repair-sensitive transition boundary varies substantially across
architectures; a principled characterization of this threshold, potentially
in terms of per-layer sparsity distribution or surviving activation variance,
would be a useful direction for future work. The present analysis is also
limited to ERK and LAMP; other allocation strategies (e.g., magnitude-based
global pruning without a structural rule, or learned allocation) may exhibit
different recoverability profiles. Comparisons against full post-training
pruning pipelines in aligned experimental settings remain to be conducted.

\section{Conclusion}

We have studied how upstream sparsity allocation affects the recoverability of
pruned neural networks under a fixed label-free repair backend. Using ASR as
the repair operator~\citep{zhan2026asr}, we show that ERK and LAMP allocations
yield substantially different post-repair accuracies at the same global
sparsity, that the direction of the advantage reverses across architectures
and datasets, and that a repair-sensitive transition regime exists in which
allocation choice matters most. Additional validations on ImageNet-100 and
DenseNet-121 confirm that the transition boundary and the appropriate repair
aggressiveness are architecture-dependent. These results suggest that
allocation and repair are not independent design choices: the structural
properties of the allocation determine the channelwise activation landscape
available for repair, and understanding this interaction is necessary for
reliably deploying label-free post-pruning recovery in practice.

\bibliographystyle{plainnat}
\bibliography{ref}

@inproceedings{evci2020rigging,
  title     = {Rigging the Lottery: Making All Tickets Winners},
  author    = {Evci, Utku and Gale, Trevor and Menick, Jacob and Castro, Pablo Samuel and Elsen, Erich},
  booktitle = {Proceedings of the 37th International Conference on Machine Learning},
  pages     = {2943--2952},
  year      = {2020},
  organization = {PMLR}
}

@inproceedings{lee2021lamp,
  title     = {{LAMP}: Extracting the Hidden Riches of Linear Anisotropic Magnitude Pruning},
  author    = {Lee, Jaeho and Park, Sejun and Mo, Sangwoo and Ahn, Sungsoo and Shin, Jinwoo},
  booktitle = {Advances in Neural Information Processing Systems},
  volume    = {34},
  pages     = {10978--10990},
  year      = {2021}
}

@inproceedings{han2015learning,
  title     = {Learning Both Weights and Connections for Efficient Neural Networks},
  author    = {Han, Song and Pool, Jeff and Tran, John and Dally, William J.},
  booktitle = {Advances in Neural Information Processing Systems},
  volume    = {28},
  year      = {2015}
}

@article{frankle2019lottery,
  title   = {The Lottery Ticket Hypothesis: Finding Sparse, Trainable Neural Networks},
  author  = {Frankle, Jonathan and Carlin, Michael},
  journal = {International Conference on Learning Representations},
  year    = {2019}
}

@inproceedings{nagel2020adaround,
  title     = {Up or Down? Adaptive Rounding for Post-Training Quantization},
  author    = {Nagel, Markus and Amjad, Rana Ali and van Baalen, Mart and Louizos, Christos and Blankevoort, Tijmen},
  booktitle = {Proceedings of the 37th International Conference on Machine Learning},
  pages     = {7197--7206},
  year      = {2020},
  organization = {PMLR}
}

@inproceedings{nagel2019dfq,
  title     = {Data-Free Quantization Through Weight Equalization and Bias Correction},
  author    = {Nagel, Markus and van Baalen, Mart and Blankevoort, Tijmen and Welling, Max},
  booktitle = {Proceedings of the IEEE/CVF International Conference on Computer Vision},
  pages     = {1325--1334},
  year      = {2019}
}

@article{lazarevich2021posttraining,
  title   = {Post-Training Deep Neural Network Pruning via Layer-Wise Calibration},
  author  = {Lazarevich, Ivan and Kozlov, Alexander and Malinin, Nikita},
  journal = {Proceedings of the IEEE/CVF International Conference on Computer Vision Workshops},
  pages   = {798--806},
  year    = {2021}
}

@inproceedings{jordan2023repair,
  title     = {{REPAIR}: {RE}normalizing Permuted Activations for Interpolation Repair},
  author    = {Jordan, Keller and Sedghi, Hanie and Saukh, Olga and Entezari, Rahim and Neyshabur, Behnam},
  booktitle = {International Conference on Learning Representations},
  year      = {2023}
}

@article{yu2026statistical,
  title   = {Statistical Energy Compensation for Post-Training {LLM} Pruning},
  author  = {Yu, Hao and others},
  journal = {arXiv preprint},
  year    = {2026}
}

@misc{zhan2026asr,
  title={Adaptive Signal Resuscitation: Channel-wise Post-Pruning Repair for Sparse Vision Networks},
  author={Qishi Zhan and Ziheng Chen and Minxuan Hu},
  year={2026},
  eprint={2605.21426},
  archivePrefix={arXiv},
  primaryClass={cs.LG},
  url={https://arxiv.org/abs/2605.21426}
}

\end{document}